\documentclass{article}

\usepackage{PRIMEarxiv}
\usepackage[utf8]{inputenc}
\usepackage[T1]{fontenc}
\usepackage{cite}
\usepackage{amsmath,amssymb,amsfonts}
\usepackage{graphicx}
\usepackage{textcomp}
\usepackage{xcolor}
\usepackage{booktabs}
\usepackage{tabularx}
\usepackage{array}
\usepackage{multirow}
\usepackage{url}
\usepackage{nicefrac}
\usepackage{microtype}
\usepackage{hyperref}
\graphicspath{{media/}}

\providecommand{\tightlist}{%
  \setlength{\itemsep}{0pt}\setlength{\parskip}{0pt}}

\def\BibTeX{{\rm B\kern-.05em{\sc i\kern-.025em b}\kern-.08em
    T\kern-.1667em\lower.7ex\hbox{E}\kern-.125emX}}

\begin{document}

\title{LoRA Fine-Tuned Models for Control Systems Course Q\&A:
A Multidimensional Evaluation of Model Scale and Rank Effects}

\author{
    Shaowen Lu \thanks{Corresponding author. Email: lusw@mail.neu.edu.cn}, Chengxu Liu, Ping Zhou and Tao Yang \\
  \textit{State Key Laboratory of Synthetical Automation for Process Industries} \\
  \textit{Northeastern University} \\
  \textit{Shenyang, China, 110819}
}


\maketitle

\begin{abstract}
Large language models (LLMs) are increasingly used to support learning in specialized university courses. However, answering questions about control systems requires a model to coordinate terminology, notation, mathematical derivations, and stepwise explanations. Responses produced directly by general-purpose models may therefore be inconsistently structured and difficult to verify. Using exercises and reference solutions from a Linear Control Systems course, we constructed a supervised fine-tuning dataset containing 360 system-user-assistant conversations. We then applied low-rank adaptation (LoRA) to Qwen2.5-3B-Instruct and Qwen2.5-7B-Instruct. Under identical data splits, inference parameters, and evaluation protocols, we compared the base and LoRA-fine-tuned models and examined LoRA ranks of $r=4$, $r=8$, and $r=16$. Evaluation used ROUGE, BERTScore, and structured-output features to assess reference-answer similarity and the stability of the Solution-Method-Teaching Points format. LoRA fine-tuning improved both reference-answer similarity and structured-output stability at both model sizes. On the current test set, 7B-r16 achieved the highest ROUGE-L (0.4093) and BERTScore-F1 (0.8643), whereas $r=8$ provided a more balanced compromise between performance and parameter efficiency. Bootstrap resampling further showed ROUGE-L gains of 0.0764 [0.0613, 0.0915] for 3B-r16 and 0.0874 [0.0687, 0.1042] for 7B-r16. Both confidence intervals were above zero, indicating stable improvements in textual similarity on the current test set. These results show that LoRA can align open-source instruction-tuned models more closely with the language and pedagogical organization of course reference answers. However, the metrics used here primarily capture textual similarity and formatting consistency. They do not directly establish improvements in domain-specific reasoning or mathematical correctness, which require further validation through expert assessment and task-specific rubrics.
\end{abstract}


\section{Introduction}\label{introduction}

Large language models have attracted substantial attention in higher education. They are increasingly used in educational question-answering systems, intelligent tutoring, and learning feedback, with applications expanding from general knowledge queries to discipline-specific instruction \cite{ref1,ref2,ref3,ref4,ref5,ref6,ref7}. Recent studies show that generative AI now supports both teaching and learning in higher education, further accelerating this transition \cite{ref2,ref8,ref9}.

Specialized university education presents distinct challenges. Students must navigate extensive bodies of knowledge and difficult theories that often involve abstract concepts, mathematical derivations, symbolic systems, and discipline-specific methods. In engineering, for example, students must integrate theoretical principles, mathematical tools, and engineering analysis. LLM-based teaching tools can address the need for on-demand questions, immediate feedback, and personalized guidance, particularly during independent study. Practical applications are already emerging across engineering education.

In computer science and software engineering, researchers have used ChatGPT and GPT-4 to solve programming problems, diagnose faulty code, and generate learning feedback. Studies have also compared their responses with feedback from human tutors. Research on teacher development and course design has examined how engineering, computing, and education instructors perceive and use generative AI. The applications include classroom teaching, curriculum design, and learning support. In university-level STEM courses, instructors have begun using generative AI for concept explanations, assignment feedback, course-material generation, and learning-activity design. These efforts show that LLMs are moving from general-purpose question answering into specific disciplinary courses. Their suitability for subjects that depend heavily on mathematical derivation and engineering explanation, such as control systems, nevertheless requires further study with specialized datasets \cite{ref10,ref11,ref12}.

Applying LLMs to specialized university courses also creates substantial challenges. A response's value depends on more than linguistic fluency. It must follow the course logic, retain essential derivation steps, and allow students to verify the explanation. The central goal of specialized course question answering is therefore not simply to generate a seemingly complete response. Instead, the model should produce structured, verifiable, and pedagogically useful answers that resemble the course reference solutions. Questions in a control systems course commonly address system modeling, differential equations, Laplace transforms, transfer functions, stability, time-domain responses, frequency-domain analysis, and controller design. Answering them requires coordination among variable definitions, physical relationships, mathematical derivations, and interpretations of the final result. A response that is fluent but lacks stable stepwise organization and teaching points may appear complete while remaining unsuitable as instructional material.

The main approaches to this problem include prompt engineering, retrieval-augmented generation (RAG), and domain-specific supervised fine-tuning \cite{ref13,ref14,ref15,ref16,ref17,ref18}. Prompt engineering can constrain response form through role definitions and output templates, whereas RAG can supply external materials such as textbooks, lecture notes, or formula sheets \cite{ref13}. These approaches primarily constrain or supplement generation at inference time. They do not directly show whether a model has learned the terminology, solution order, and pedagogical style of course reference answers. By adjusting the model's output distribution with target-task examples \cite{ref14,ref15,ref18}, domain-specific supervised fine-tuning is better suited to testing whether a model can reproduce stable course-specific response patterns.

Prompt engineering and RAG primarily operate at inference time and do not directly alter the model parameters. Domain-specific supervised fine-tuning can modify the output distribution, but specialized course datasets in higher engineering education are often small and computational resources are limited. Full-parameter fine-tuning is therefore not always economical \cite{ref19,ref20,ref21}. Low-rank adaptation (LoRA) \cite{ref22} is a parameter-efficient fine-tuning method. It freezes the pretrained weights and represents task-specific updates as the product of two low-rank matrices. This representation adapts the model to a domain distribution with relatively few trainable parameters. LoRA thus provides a parameter-efficient bridge between retaining the general capabilities of a base model and fitting the response patterns of course-specific data. This property makes model customization feasible with small course datasets and limited hardware.

LoRA has recently been applied to model customization in education because it enables domain adaptation under constrained computing resources.

The AI-University platform, for example, used a graduate finite-element course as a case study. It constructed training data from course videos, lecture notes, and textbooks and combined LoRA fine-tuning with RAG to align responses with a specific instructor's materials \cite{ref23}. In statistics education, researchers compared zero-shot prompting, few-shot prompting, and LoRA-based supervised fine-tuning for generating feedback on student answers, examining both accuracy and pedagogical value \cite{ref24}. For automated writing evaluation, AiAWE used a LoRA-adapted, instruction-tuned model to score student argumentative essays. The study demonstrated the feasibility of building an open evaluation model from a small educational dataset \cite{ref25}. Together, these studies indicate that LoRA provides a practical route to model customization for specific courses, tasks, and assessment criteria.

Previous studies have applied LoRA to tasks such as mathematical problem solving, programming support, and standardized-test question answering. Their findings support its ability to learn discipline-specific response patterns at low parameter cost while retaining the general capabilities of the base model. Most applications, however, treat LoRA as a fixed fine-tuning tool and focus on its resource savings relative to full-parameter fine-tuning. Less attention has been paid to the joint effects of adaptation parameters, such as LoRA rank, and base-model size. In an engineering course such as control systems, responses depend heavily on consistent terminology, coherent derivations, and standardized pedagogical formatting. Base-model size may constrain general semantic understanding and foundational reasoning, whereas LoRA rank determines the capacity of the adaptation module to fit the course-specific distribution. Existing research has not systematically examined how model size and LoRA rank jointly affect course question answering. It remains unclear whether a higher rank can compensate for the limitations of a smaller model or whether increasing rank produces diminishing returns in a larger model. This gap makes it difficult for practitioners to select models and adaptation settings that match their hardware constraints and instructional goals.

Using Linear Control Systems as a case study, this study investigates how model size and LoRA rank jointly affect adaptation for course question answering. The objective is not merely to determine whether LoRA works as a fixed fine-tuning tool, but to examine the practical trade-off between parameter efficiency and performance. The study addresses four research questions:

\begin{enumerate}
\def\labelenumi{\arabic{enumi}.}
\tightlist
\item
  Does LoRA fine-tuning on control systems course data improve the lexical and semantic similarity between model-generated responses and reference answers across different model sizes?
\item
  Can LoRA fine-tuning improve structured instructional output by making the Solution-Method-Teaching Points format more stable?
\item
  How do model size and LoRA rank affect adaptation performance, parameter efficiency, and marginal returns?
\item
  How should these results be interpreted without equating similarity and formatting metrics with domain-specific reasoning ability?
\end{enumerate}

This study makes three contributions.

\begin{enumerate}
\def\labelenumi{\arabic{enumi}.}
\tightlist
\item
  \textbf{Dataset:} We constructed a structured question-answering dataset for \emph{Linear Control Systems}, a third-year core course in automation. The dataset contains 360 samples, each comprising a question, additional conditions, and a reference solution. All source data were manually reviewed. The dataset supports supervised fine-tuning experiments for a course-specific question-answering model.
\item
  \textbf{Model-performance experiments:} We systematically examined the joint effects and marginal returns of model size and LoRA rank in course question-answering adaptation. Under a unified experimental protocol, Qwen2.5-3B and Qwen2.5-7B were fine-tuned with LoRA ranks of $r=4$, $r=8$, and $r=16$. The experiments characterized how base models of different sizes respond to adaptation capacity, how gains change with rank, and how parameter efficiency decreases as rank increases. These results provide quantitative guidance for selecting course-model configurations under resource constraints.
\item
  \textbf{Evaluation framework:} We compared base and LoRA-fine-tuned models using ROUGE, BERTScore, structured-output features, and bootstrap confidence intervals. The resulting framework focuses on reference-answer similarity, the organization of instructional responses, and the stability of observed improvements. We explicitly distinguish textual similarity and structural stability from domain-specific correctness. Formula accuracy, derivational rigor, and instructional reliability remain questions for future expert evaluation.
\end{enumerate}

\section{Related Work}\label{related-work}

\subsection{Adaptation Approaches for Specialized Course Question Answering}\label{adaptation-approaches-for-specialized-course-question-answering}

Prompt engineering, retrieval-augmented generation, and domain-specific supervised fine-tuning are common approaches to adapting models for specialized course question answering \cite{ref13,ref14,ref15,ref16,ref17,ref18}. Prompt engineering uses role definitions, task constraints, and output templates to guide the model toward a particular response format. It is inexpensive to implement and allows rapid adjustment of response style. In chat-style data, the system message serves as a task-level prompt that specifies the model's role, domain, and response requirements, while the user message contains the question. Prompt engineering acts primarily at inference time, and its effectiveness depends on prompt design and the model's existing capabilities. Because it does not change model parameters, it cannot ensure stable learning of the terminology, step order, and stylistic conventions found in course reference answers.

RAG introduces external knowledge, such as textbooks, lecture notes, formula sheets, or related documents, before generation \cite{ref13,ref17}. This approach can compensate for the absence of course materials in the model's parameters and can improve response traceability. Its performance nevertheless depends on retrieval quality, document segmentation, context length, and the generator's ability to integrate retrieved evidence. For control systems questions with mathematical derivations and structured procedures, external material alone does not ensure alignment with the organization of the course reference answer. RAG evaluation must also assess faithfulness to the retrieved documents, making it different from conventional text-generation evaluation.

Domain-specific supervised fine-tuning directly adjusts the model's generation distribution through target-task examples \cite{ref14,ref15}. It exposes the model to the terminology, derivation order, and pedagogical style of course answers. Compared with prompt-based methods that impose constraints only during inference, fine-tuning is better suited to testing whether a model can learn stable response patterns from a specific course. However, full-parameter fine-tuning on a small course dataset creates risks of overfitting and catastrophic forgetting and generally requires substantial computational resources. Parameter-efficient fine-tuning (PEFT), which updates only a small set of added or low-rank parameters, is more appropriate for small datasets and constrained hardware \cite{ref19,ref20,ref21,ref22,ref26,ref27}. These three approaches are not mutually exclusive. Fine-tuning can internalize course-specific terminology and stylistic conventions, while RAG can supply traceable external knowledge during generation. In this study, we use LoRA as the principal adaptation method and examine whether it improves reference-answer similarity and the stability of structured responses.

\subsection{Parameter-Efficient Fine-Tuning and Low-Rank Adaptation}\label{parameter-efficient-fine-tuning-and-low-rank-adaptation}

As LLMs have grown, the computational and memory costs of full-parameter fine-tuning have become major barriers to model customization \cite{ref19,ref20,ref21,ref22,ref26,ref27}. PEFT freezes most pretrained parameters and introduces a small number of trainable parameters for task adaptation \cite{ref22,ref26,ref27}. These methods lower the cost of domain-specific customization and make adaptation experiments feasible in resource-constrained settings. For course question answering, PEFT is attractive because it can learn course-specific response structures, terminology, and solution styles at relatively low training cost.

Existing PEFT methods include adapters, prompt tuning, and LoRA \cite{ref20,ref21,ref22,ref26,ref27}. They differ in parameter count, training stability, deployment complexity, and inference overhead. Compared with updating all model parameters, PEFT is better suited to small domain datasets and hardware-constrained environments. It also allows multiple task-specific modules to be maintained for a single base model. PEFT therefore provides a practical engineering approach to building course-level question-answering models.

LoRA is a representative PEFT method. It freezes the pretrained weights and uses low-rank matrices to represent the weight update required for task adaptation. For a pretrained weight matrix \((W_{0}\in \mathbb{R}^{d\times k})\), LoRA does not update \((W_{0})\) directly. Instead, it expresses the update as the product of two low-rank matrices:

\[
\Delta W=BA, \quad B\in \mathbb{R}^{d\times r}, \quad A\in \mathbb{R}^{r\times k}, \quad r\ll \min(d,k)
\]

The output of a LoRA-augmented linear transformation is:

\[
h=W_{0}x+\frac{\alpha }{r}BAx
\]

Here, A and B are trainable low-rank matrices, r is the LoRA rank, and \(\alpha /r\) is the scaling factor. During training, the original weight \(W_{0}\) remains frozen, and only A and B are updated. This design substantially reduces the number of trainable parameters.

LoRA's modular design makes it compatible with different Transformer architectures and allows multiple task adapters to be attached to the same base model \cite{ref19,ref20,ref21,ref22}. This property is particularly useful in education. Each course or semester can have an independent LoRA adapter attached to a shared base model, eliminating the need to deploy a separate full model. Adapter files are typically only tens to hundreds of megabytes, which facilitates storage, distribution, and version control. LoRA is therefore well suited to the practical requirement of one adapter per course in university teaching.

The rank r is a key hyperparameter because it determines the dimensions and representational capacity of the low-rank matrices. A smaller r improves parameter efficiency, whereas a larger r may improve task fit but also increase the risk of overfitting. Previous studies of question answering and instruction following have commonly used r values from 4 to 64 \cite{ref22}. Performance in small-data settings does not necessarily increase linearly with parameter count. Selecting the LoRA rank therefore requires a trade-off among parameter efficiency, adaptation capacity, and generalization risk. We compare $r=4$, $r=8$, and $r=16$ to represent low, medium, and high adaptation capacities. Results from Qwen2.5-3B and Qwen2.5-7B are then used to examine performance saturation and the effects of model size.

\subsection{Evaluation of Educational Question-Answering Models}\label{evaluation-of-educational-question-answering-models}

Educational question-answering models cannot be evaluated solely by response fluency. In a control systems course, answer quality also depends on accurate terminology, clear variable definitions, coherent formula derivations, verifiable conclusions, and pedagogical value. Automatic text metrics can quantify similarity to a reference answer, but they cannot independently establish reliable domain-specific reasoning.

ROUGE measures lexical overlap between generated and reference answers \cite{ref28}, whereas BERTScore captures semantic similarity \cite{ref29}. These metrics can indicate whether fine-tuning brings model responses closer to the language of course reference answers. Control systems solutions, however, often contain symbols, formulas, and multistep derivations. Neither lexical nor semantic similarity directly validates mathematical correctness \cite{ref30}. A response may resemble the reference answer while containing an incorrect formula, a skipped derivation step, or an unstated assumption.

In addition to ROUGE and BERTScore, we therefore measure structured-output features, including Solution, Method, and Teaching Points. These features indicate whether the model consistently produces the organizational structure expected in an instructional setting. This design reflects the scope of the study. We evaluate the effects of LoRA on reference-answer similarity and the stability of pedagogical structure without treating these metrics as direct measures of domain-specific correctness or reasoning ability.

Teachability, defined here as the extent to which an answer helps students understand, follow, and verify a solution, generally requires a human-evaluation rubric. Existing human evaluations of educational question answering often assess correctness, completeness, clarity of derivation steps, adequacy of variable explanations, and communication of the conditions under which a conclusion applies \cite{ref31,ref32}. These dimensions extend beyond automatic text metrics. Expert assessment or task-specific rubrics are therefore still needed to evaluate formula accuracy, derivational rigor, and instructional usefulness.

In summary, previous work has established foundations for educational question answering, model adaptation, and parameter-efficient fine-tuning. Yet the relationships among model size, LoRA rank, reference-answer similarity, and structured instructional output remain underexplored for small engineering-course datasets. This study addresses that gap. It also distinguishes changes in text and format from evidence of domain-specific reasoning, providing an explicit boundary for future discussions of LLM evaluation in engineering education.

\section{Linear Control Systems Question-Answering Dataset}\label{linear-control-systems-question-answering-dataset}

\subsection{Course Context of the Source Data}\label{course-context-of-the-source-data}

\emph{Linear Control Systems} is a core upper-level undergraduate course in automation, electrical engineering, and related disciplines. It also provides an important foundation for subsequent courses in modern control theory, robotics, and industrial automation. The curriculum covers dynamic-system modeling, Laplace transforms, transfer functions, time- and frequency-domain analysis, stability assessment, root-locus analysis, and controller design. It therefore combines mathematical derivation with engineering analysis.

Course instruction and assessment require students not only to obtain correct results but also to present stepwise derivations, consistent notation, and interpretations of engineering implications. A complete solution typically includes the modeling process, key equations, intermediate calculations, and the meaning of the result. Its clear problem-derivation-explanation structure makes it suitable for constructing a supervised fine-tuning dataset for instructional question answering. The source questions covered dynamic-system modeling, transfer-function models, feedback analysis, state-space models and linearization, controller design, frequency-domain analysis, performance analysis, and robustness analysis. Based on their task characteristics, we grouped them into computational derivation, analytical discussion, control-system design, and conceptual understanding questions. These categories provided the basis for constructing samples with both technical and pedagogical content.

\subsection{Data Collection and Preprocessing}\label{data-collection-and-preprocessing}

We selected questions with reference to the textbook chapter structure and course syllabus to improve coverage of core knowledge and ensure that answers were standardized and verifiable. Questions were drawn from textbook exercises and online problem sets. We prioritized representative exercises that required modeling, analysis, or design. Composite problems with multiple subquestions were divided into relatively independent question-answer samples according to their knowledge points and solution objectives.

Questions that required only a single formula substitution, had excessively short solutions, or were too open-ended to support a consistent reference answer were excluded. After selection, decomposition, and standardization, the final dataset comprised 360 question-answer samples from the Linear Control Systems course.

The original exercises varied in wording, variable naming, and information organization. We therefore standardized them before constructing the dataset. Mathematical notation, variable names, and system representations were harmonized. Transfer functions, input and output signals, system parameters, and state variables followed consistent naming conventions, reducing noise caused by alternative expressions for the same concept.

For questions that depended on block diagrams, signal-flow graphs, circuit diagrams, or mechanical schematics, we converted the visual information into structured textual descriptions. The descriptions specified component connections, feedback paths, and system parameters, allowing the model to understand each question without visual input. Initial states, boundary conditions, parameter ranges, and performance requirements were also retained explicitly to support the reference derivations.

\subsection{Standardization and Structured Construction of Reference Answers}\label{standardization-and-structured-construction-of-reference-answers}

After standardizing the questions, we reconstructed the reference solutions into a consistent format. Textbook answers varied in derivational detail and explanatory depth. Using the original reference solutions and course teaching requirements, we rewrote each assistant response under the main heading Solution, with Method and Teaching Points as subsections.

The Method subsection presents the choice of method, variable definitions, formula derivations, intermediate calculations, parameter substitution, and final result. The Teaching Points subsection summarizes key concepts, common errors, physical meanings, and engineering interpretations. This structure provides an answer while retaining the derivation path and pedagogical guidance needed for classroom explanation.

The initial dataset was manually reviewed for accuracy and instructional reliability. The review focused on formula correctness, completeness of calculations, consistency of notation, LaTeX formatting, and the accuracy of teaching points. Approximately 20\% of all samples were spot-checked, and identified issues were revised in accordance with the review comments. This process improved dataset standardization and usability.

\subsection{Conversion to a Dialogue Format for Model Fine-Tuning}\label{conversion-to-a-dialogue-format-for-model-fine-tuning}

After question selection, standardization, and reference-answer reconstruction, we converted the data into chat-style JSONL for supervised LLM fine-tuning. Each sample comprises system, user, and assistant messages.

\textbf{System message.} All 360 samples use the same system message. It assigns the model the role of a teaching assistant for the Linear Control Systems course and requires standard control-theory terminology and LaTeX notation. It also asks the model to summarize key teaching points at the end of each answer. A uniform system message ensures consistent role definitions and stylistic constraints across samples and avoids additional noise from variation in system prompts.

\textbf{User message.} The user message is not a plain question. It follows a four-part structure. (1) \textbf{Problem Description} states the context, known conditions, and objective while preserving the physical setting and engineering parameters. (2) \textbf{Subproblems} divides the solution into four to six steps, guides staged reasoning, and provides the model with a clear reference path. (3) \textbf{Additional Information} supplies physical parameters, mathematical models, and key formulas, allowing the problem to be solved without textbook context. (4) \textbf{Constraints} specifies response requirements and boundary conditions, including LaTeX formatting, independent treatment of the problem, and explicit conclusions. This structure preserves the modeling, derivational, and interpretive content of control systems problems. It also reduces the variability of open-ended question answering through stepwise guidance and explicit constraints, making the data better suited to supervised learning from paired inputs and outputs.

\textbf{Assistant message.} The assistant message contains the structured reference answer in the three-level Solution-Method-Teaching Points format described in Section~\ref{standardization-and-structured-construction-of-reference-answers}. Method presents the complete derivation and calculation, while Teaching Points summarizes key concepts and common errors.

This format is directly compatible with supervised fine-tuning frameworks for major open-source LLMs, including Llama, Qwen, and DeepSeek.

The final dataset contains 360 question-answer samples. To maintain similar distributions of knowledge modules and question types across subsets, we used stratified random sampling based on knowledge point and question category. The data were split into training, validation, and test sets, as summarized in Table~\ref{tab:dataset-split}.

\begin{table}[t]
\centering
\caption{Dataset split for the control systems course question-answering dataset}
\label{tab:dataset-split}
\scriptsize

\resizebox{\textwidth}{!}{%
\begin{tabular}{@{}llll@{}}
\toprule
\textbf{Data subset} & \textbf{Number of samples} & \textbf{Proportion} & \textbf{Primary purpose} \\
\midrule
Training set & 270 & 75\% & Parameter learning during LoRA fine-tuning \\
Validation set & 36 & 10\% & Model selection and hyperparameter tuning \\
Test set & 54 & 15\% & Evaluation of generalization to unseen questions \\
 &  &  &  \\
\bottomrule
\end{tabular}%
}
\end{table}

After splitting, we checked each subset for exact duplicates at the question-text, reference-answer, and complete-dialogue levels. No exact duplicates were found, reducing the risk that data leakage affected model evaluation.

\subsection{Statistical Analysis and Summary of Dataset Characteristics}\label{statistical-analysis-and-summary-of-dataset-characteristics}

\subsubsection{Basic Statistical Characteristics}\label{basic-statistical-characteristics}

The dataset contains 360 structured question-answer samples. It is written in English, consistent with the language of the source textbooks, and covers the core topics of a Linear Control Systems course:

\begin{itemize}
\tightlist
\item
  \textbf{Dynamic-system modeling:} mass-spring-damper systems, rotational machinery and belt drives, direct-current motors, thermodynamic systems, RLC circuits, liquid-level systems, satellite orbital dynamics, and other physical domains. These questions train models to construct differential-equation models in different engineering contexts;
\item
  \textbf{Laplace transforms and transfer functions:} solutions of ordinary differential equations using Laplace transforms, transfer-function derivation, and representation and reduction of block diagrams;
\item
  \textbf{Time-domain analysis and stability:} zero-input and zero-state responses, step responses and transient-performance measures, stability assessment from pole locations, and classification of equilibria;
\item
  \textbf{Linearization and state space:} Jacobian linearization of nonlinear systems at equilibrium points, state-space representations, and controllability and observability analysis;
\item
  \textbf{Frequency-domain analysis:} construction and interpretation of Bode plots, the Nyquist stability criterion, frequency response, and bandwidth analysis;
\item
  \textbf{Controller design:} parameter tuning for proportional (P), proportional-derivative (PD), and proportional-integral (PI) controllers, root-locus design, and disturbance rejection using integral controllers.
\end{itemize}

The samples fall into four categories according to task characteristics. Computational derivation questions, such as transfer-function calculation, ordinary differential equation solving, and linearization, form the largest category and reflect the mathematically oriented nature of the course. Analytical discussion questions require qualitative interpretation of results, including stability judgments, response behavior, and physical meaning. Control-system design questions, such as controller-parameter selection and root-locus design, simulate engineering design decisions. Conceptual understanding questions assess deeper understanding of control-theory concepts, such as the physical interpretation of controllability and the implications of modeling assumptions. Stratified sampling maintained similar proportions of the four categories across the training, validation, and test sets, supporting comparisons across question types.

Control systems questions frequently involve transfer-function derivation, stability analysis, and controller design. The reference answers therefore contain many mathematical formulas, variable symbols, and LaTeX expressions. Each assistant response contains approximately 15-25 LaTeX environments on average, including inline and display mathematics. This feature reflects the strong dependence of control systems question answering on mathematical expression. It can help a model learn specialized terminology, mathematical conventions, and engineering solution procedures, while placing substantial demands on notation consistency and long, multistep derivations.

\subsubsection{Coverage of Answer Structures}\label{coverage-of-answer-structures}

Following the structured design described in Section~\ref{standardization-and-structured-construction-of-reference-answers}, all 360 reference answers contain the three components Solution, Method, and Teaching Points. Each structural marker therefore has 100\% coverage. Method contains the complete derivation and calculation, while Teaching Points summarizes key concepts and common errors in four to five bullet points. The uniform structure reduces formatting variation and can stabilize model outputs during supervised fine-tuning. It also provides an analytical dimension beyond textual similarity, namely whether the generated response exhibits the complete Solution-Method-Teaching Points organization.

In summary, the dataset covers the principal knowledge modules and question types of the Linear Control Systems course. It also standardizes the full pipeline from question wording to reference-answer structure. Table~\ref{tab:3-1} summarizes the dataset.

\begin{table}[t]
\centering
\caption{Basic information about the control systems course question-answering dataset}
\label{tab:3-1}
\small
\begin{tabularx}{\linewidth}{@{}p{0.30\linewidth}X@{}}
\toprule
\textbf{Item} & \textbf{Description} \\
\midrule
Data sources & Exercises and reference solutions from Linear Control Systems textbooks \\
Language & English \\
Data format & JSONL (chat-style) \\
Sample structure & System-user-assistant dialogue; four-part user message (Problem Description / Subproblems / Additional Information / Constraints); three-level assistant response (Solution-Method-Teaching Points) \\
Total samples & 360 \\
Training set & 270 (75\%) \\
Validation set & 36 (10\%) \\
Test set & 54 (15\%) \\
Splitting method & Stratified random sampling by knowledge point and question type \\
Exact duplicate check across subsets & No duplicate user messages, assistant responses, or complete samples \\
\bottomrule
\end{tabularx}
\end{table}

This procedure produced a supervised fine-tuning dataset for instructional question answering in control systems. The dataset preserves the course's technical and derivational features. Its uniform dialogue format, four-part question structure, and pedagogically structured answers provide a consistent basis for LoRA fine-tuning across model sizes.

\section{Methods}\label{methods}

\subsection{Research Framework and Experimental Variables}\label{research-framework-and-experimental-variables}

\begin{figure}[t]
    \centering
    \includegraphics[width=\linewidth]{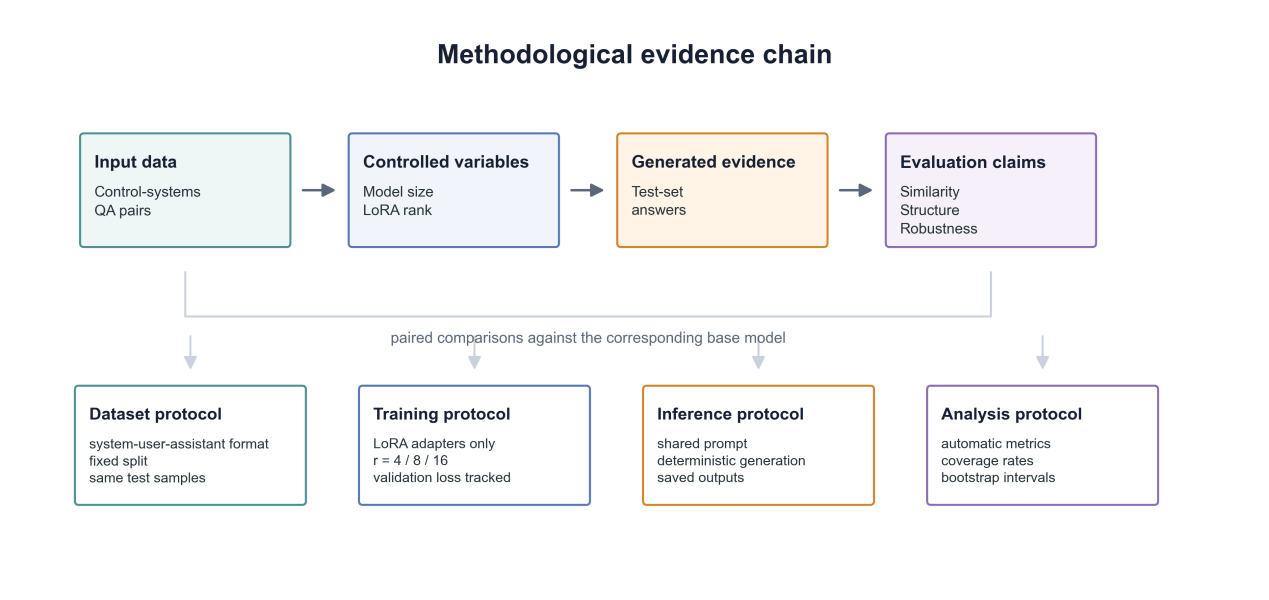}
    \caption{Overall research workflow}
    \label{fig:workflow}
\end{figure}

We developed a framework for adapting and evaluating models for question answering in a control systems course. Using course question-answer data, the framework converts textbook exercises and reference solutions into system-user-assistant dialogue samples. It then applies LoRA-based supervised fine-tuning to adapt open-source instruction-tuned models to the course domain. The overall workflow comprises dataset construction, dialogue-format conversion, base-model selection, LoRA fine-tuning, test-set inference, and multidimensional evaluation. Fig.~\ref{fig:workflow} presents the complete research workflow.

Unlike studies that examine the fine-tuning of a single model, our experiments vary two factors. First, we compare Qwen2.5-3B-Instruct and Qwen2.5-7B-Instruct to examine how smaller and larger models adapt to the same course data. Second, we compare LoRA ranks of $r=4$, $r=8$, and $r=16$ at each model size to assess how adaptation capacity affects reference-answer similarity and structured instructional output. The training data, validation and test splits, target modules, learning rate, number of training epochs, inference parameters, and evaluation metrics remain constant. Only base-model size and LoRA rank vary.

\subsection{Base Models and Experimental Environment}\label{base-models-and-experimental-environment}

We selected Qwen2.5-3B-Instruct and Qwen2.5-7B-Instruct as the two base models \cite{ref33}. Both are instruction-tuned models designed for dialogue generation and instruction following. The 3B model was used to test whether a smaller model could learn the structure of course reference answers through LoRA under resource constraints. The 7B model was used to examine whether greater model size improved reference-answer similarity and structured-output stability. Both models used identical data splits, LoRA rank settings, and evaluation protocols, supporting an interpretable comparison of model size. Table~\ref{tab:4-2} summarizes the base models and their experimental roles.

\begin{table}[t]
\centering
\caption{Base models and experimental purposes}
\label{tab:4-2}
\scriptsize
\resizebox{\textwidth}{!}{%
\begin{tabular}{@{}lll@{}}
\toprule
\textbf{Model} & \textbf{Experimental role} & \textbf{Purpose of comparison} \\
\midrule
Qwen2.5-3B-Instruct & Smaller instruction-tuned model & Evaluate the feasibility of LoRA course adaptation under resource constraints \\
Qwen2.5-7B-Instruct & Larger instruction-tuned model & Evaluate the effects of greater model size on answer similarity and structured output \\
\bottomrule
\end{tabular}%
}
\end{table}

All experiments, including model loading, LoRA fine-tuning, inference, and evaluation, were conducted in the same Linux server environment. The server contained one Tesla V100-DGXS-32GB GPU with 32 GB of memory. It ran Linux with Ubuntu kernel 5.15.0-177-generic and CUDA 12.1. This environment supported parameter-efficient fine-tuning of both the 3B and 7B models.

The software environment included Python 3.10.20, PyTorch 2.5.1+cu121, Transformers 5.8.0, PEFT 0.19.1, and TRL 1.4.0. Transformers was used for model loading, tokenization, and generation; PEFT implemented LoRA adaptation; and TRL supported supervised fine-tuning. We fixed the random seed and recorded the main dependency versions and runtime settings to improve reproducibility.

\subsection{LoRA-Based Supervised Fine-Tuning}\label{lora-based-supervised-fine-tuning}

We used LoRA for parameter-efficient adaptation within a supervised fine-tuning framework \cite{ref20,ref21,ref22}. LoRA freezes the original base-model weights and introduces trainable low-rank matrices into specified linear projection layers. For a pretrained weight matrix W, LoRA represents the update as \(\Delta W=BA\), where r determines the dimensions of the low-rank matrices. Only A and B are updated during training; the original W remains frozen. This design reduces both the trainable parameter count and memory requirements while allowing the model to learn course-specific terminology, derivation order, and pedagogical response patterns.

Given an input prompt \(x\) and target response \(y=(y_{1},y_{2},\ldots ,y_{T})\), supervised fine-tuning updates the LoRA adapter parameters by minimizing the negative log-likelihood of the target tokens:

\[
\mathcal{L}_{\mathrm{SFT}}=-\sum_{t=1}^{T}\log p_{\theta}(y_{t}\mid x,y_{<t})
\]

Here, \(p_{\theta}(y_{t}\mid x,y_{<t})\) is the probability that the model assigns to the \(t\)th target token given the input prompt and preceding target tokens.

To ensure comparability, all experimental groups used identical training data, epochs, learning rates, batch sizes, maximum sequence lengths, target modules, and random seeds. The principal variables were base-model size and LoRA rank r. LoRA alpha (\(\alpha\)) was set to \(\alpha=2r\). The target modules were the \texttt{q\_proj}, \texttt{k\_proj}, \texttt{v\_proj}, and \texttt{o\_proj} projections in Transformer self-attention \cite{ref34}. We compared $r=4$, $r=8$, and $r=16$ to examine how low-rank dimensionality affected adaptation to course question answering. Table~\ref{tab:4-3} lists the training settings.

\begin{table}[t]
\centering
\caption{Parameters for LoRA-based supervised fine-tuning}
\label{tab:4-3}
\small
\begin{tabularx}{\linewidth}{@{}p{0.30\linewidth}X@{}}
\toprule
\textbf{Parameter} & \textbf{Setting} \\
\midrule
Base models & Qwen2.5-3B-Instruct; Qwen2.5-7B-Instruct \\
Fine-tuning method & LoRA (SFT) \\
Training epochs & 5 \\
Maximum sequence length & 2048 \\
Training batch size & 1 \\
Evaluation batch size & 1 \\
Gradient-accumulation steps & 8 \\
Effective batch size & 8 \\
Learning rate & 5e-5 \\
Warm-up ratio & 0.03 \\
Weight decay & 0.01 \\
LoRA rank r & 4/8/16 \\
LoRA scaling factor alpha & 8/16/32 \\
LoRA dropout & 0.05 \\
Target modules & q\_proj, k\_proj, v\_proj, o\_proj \\
Random seed & 42 \\
Deterministic training & True \\
Training precision & auto, resolved to torch.bfloat16 \\
Best-model selection & Minimum eval\_loss \\
3B LoRA adapter parameters & $r=4$: 1,843,200; $r=8$: 3,686,400; $r=16$: 7,372,800 \\
7B LoRA adapter parameters & $r=4$: 2,523,136; $r=8$: 5,046,272; $r=16$: 10,092,544 \\
Learning-rate scheduler & cosine \\
Evaluation/save strategy & Evaluate and save after each epoch; retain at most two checkpoints \\
Packing & False \\
LoRA bias & none \\
\bottomrule
\end{tabularx}
\end{table}

After training, we saved the LoRA adapter, tokenizer, training configuration, and evaluation configuration for each experimental group. We did not merge the base-model and LoRA weights. During inference, each base model was loaded and its corresponding adapter attached. This approach preserved the adapters for each rank and model size, facilitating comparison on a common test set.

\subsection{Inference, Generation, and Evaluation Metrics}\label{inference-generation-and-evaluation-metrics}

We evaluated performance on control systems question answering with a framework that combined automatic metrics and structured-output features. The framework assessed both the similarity of generated responses to reference answers and their conformity to the expected pedagogical organization. The evaluated systems comprised the 3B base model, the 3B LoRA-fine-tuned models, the 7B base model, and the 7B LoRA-fine-tuned models. Each fine-tuned model was evaluated at $r=4$, $r=8$, and $r=16$.

During inference, the base and LoRA-fine-tuned models used the same test set, system prompt, and generation parameters. The test set contained 54 samples. The generation limit, \texttt{max\_new\_tokens}, was 2048. We set \texttt{do\_sample=False} for deterministic decoding. Inference precision was set to \texttt{auto} and resolved to bfloat16, and the random seed was fixed at 42. Although \texttt{temperature=0.2} and \texttt{top\_p=0.9} were recorded in the runtime configuration, they did not affect decoding because sampling was disabled.

First, we used ROUGE to measure textual similarity between generated responses and reference answers \cite{ref28}. ROUGE-1, ROUGE-2, and ROUGE-L quantify surface-level similarity through unigram overlap, bigram overlap, and the longest common subsequence, respectively. We then used BERTScore-F1 to measure semantic similarity \cite{ref29}. Because the dataset is in English, \texttt{bert-base-uncased} served as the BERTScore encoder, complementing ROUGE with a semantic comparison. ROUGE-1, ROUGE-2, and ROUGE-L were calculated as F1 scores with \texttt{use\_stemmer=True}. BERTScore used \texttt{bert-base-uncased} with \texttt{num\_layers=12}, \texttt{batch\_size=2}, and no baseline rescaling.

Because instructional question answering in control systems requires clear pedagogical organization in addition to similarity with the reference answer, we defined structured-output features as auxiliary metrics. The assistant responses in the fine-tuning data use the Solution-Method-Teaching Points structure. We therefore used \texttt{has\_solution}, \texttt{has\_method}, and \texttt{has\_teaching\_points} to detect whether generated responses contained an overall solution, a method description, and a summary of teaching points. Their test-set coverage was reported as \texttt{has\_solution\_rate}, \texttt{has\_method\_rate}, and \texttt{has\_teaching\_points\_rate}. These features describe pedagogical formatting and should not be interpreted as direct measures of response quality. Table~\ref{tab:4-4} summarizes the evaluation metrics.

For any structural marker \((c)\), its coverage rate on the test set is:

\[
\mathrm{Rate}(c)=\frac{1}{N}\sum_{i=1}^{N}\mathbb{I}(c\in \hat{y}_{i})
\]

Here, \(c\in \{Solution,Method,Teaching Points\}\), and \(N\) is the number of test samples. The model-generated response for sample \(i\) is \(\hat{y}_{i}\), and \(\mathbb{I}(\cdot)\) is the indicator function.

\begin{table}[t]
\centering
\caption{Evaluation metrics for the control systems course question-answering models}
\label{tab:4-4}
\scriptsize
\resizebox{\textwidth}{!}{%
\begin{tabular}{@{}lll@{}}
\toprule
\textbf{Metric category} & \textbf{Metric} & \textbf{Measured property} \\
\midrule
Textual similarity & ROUGE-1, ROUGE-2, ROUGE-L & Lexical overlap between generated responses and reference answers \\
Semantic similarity & BERTScore-F1 & Semantic proximity between generated responses and reference answers \\
Structured output & has\_solution, has\_method, has\_teaching\_points & Whether a response follows the pedagogical organization \\
Structured-output rate & has\_solution\_rate, has\_method\_rate, has\_teaching\_points\_rate & Proportion of test samples exhibiting each structural feature \\
\bottomrule
\end{tabular}%
}
\end{table}

The evaluation was designed to measure reference-answer-oriented generation, not to certify the models' domain-specific reasoning quality \cite{ref30,ref31,ref32,ref35,ref36}. ROUGE and BERTScore primarily capture lexical and semantic proximity, while the structured-output metrics measure template stability. None can replace expert assessment of formula accuracy, derivational rigor, or instructional usefulness.

\subsection{Bootstrap Robustness Analysis}\label{bootstrap-robustness-analysis}

To assess the stability of test-set performance, we used bootstrap resampling to calculate 95\% confidence intervals for the principal metrics \cite{ref37,ref38}. For each iteration, 54 samples were drawn with replacement from the 54-item test set, and the mean metric was recalculated. This procedure was repeated B=10,000 times. The 2.5th and 97.5th percentiles of the resampled distribution defined the 95\% confidence interval.

We used a paired bootstrap to estimate improvements of fine-tuned models over their corresponding base models. Metric differences were computed on identical sample indices, allowing us to estimate how variation in test-set composition affected the measured gains. The result files did not retain per-sample BERTScore values, so no bootstrap confidence interval was calculated for BERTScore-F1. Instead, BERTScore-F1 is reported only as an overall mean and serves as a supplementary semantic-similarity metric. The bootstrap robustness analysis therefore focuses on ROUGE and structured-output metrics.

For test sample \(i\), the paired improvement of a fine-tuned model over its base model is:

\[
\Delta m_{i}=m_{i}^{\mathrm{FT}}-m_{i}^{\mathrm{Base}}
\]

Here, \(m_{i}^{\mathrm{FT}}\) and \(m_{i}^{\mathrm{Base}}\) are the metric values of the fine-tuned and base models, respectively, on test sample \(i\). In bootstrap iteration \(b\), the mean improvement is:

\[
\bar{\Delta}^{(b)}=\frac{1}{N}\sum_{i\in S^{(b)}}\Delta m_{i}
\]

After B=10,000 iterations, the 95\% confidence interval is:

\[
\mathrm{CI}_{95\%}=\Bigl[\,Q_{2.5}\bigl(\{\bar{\Delta}^{(b)}\}_{b=1}^{B}\bigr),\; Q_{97.5}\bigl(\{\bar{\Delta}^{(b)}\}_{b=1}^{B}\bigr)\Bigr]
\]

Here, \((Q_{2.5})\) and \((Q_{97.5})\) are the 2.5th and 97.5th percentiles of the resampled distribution. If the entire 95\% confidence interval for a metric improvement is above zero, the improvement is considered stable under resampling. An interval that crosses zero indicates substantial uncertainty in that improvement.

\subsection{Experimental Controls and Reproducibility}\label{experimental-controls-and-reproducibility}

We used a common experimental-control strategy to ensure fair comparisons. All models shared the same training, validation, and test splits. All inference runs used the same system prompt and user problem. The base and LoRA-fine-tuned models generated responses for the same test samples, and the same evaluation script calculated all automatic and structured-output metrics.

The random seed was fixed at 42 during both training and inference, and deterministic decoding with \texttt{do\_sample=False} was used for inference. During training, each model was evaluated on the validation set and saved after every epoch. The checkpoint with the lowest \texttt{eval\_loss} was selected for test-set evaluation.

These controls allow observed differences to be attributed primarily to base-model size and LoRA rank. However, the experiments used only one random seed and a 54-item test set. The results should therefore be interpreted as observations under the current dataset and runtime configuration. We restrict our conclusions to changes in reference-answer similarity and structured instructional output and do not treat them as evidence of general improvements in domain-specific reasoning.

\section{Experimental Results and Analysis}\label{experimental-results-and-analysis}

\subsection{Experimental Setting and Overview of Results}\label{experimental-setting-and-overview-of-results}

Under the unified protocol described in Section~\ref{methods}, this section reports the results of LoRA fine-tuning Qwen2.5-3B-Instruct and Qwen2.5-7B-Instruct for question answering in a control systems course. The experiments compare two dimensions: base-model size (3B and 7B) and LoRA rank ($r=4$, $r=8$, and $r=16$). All models used the same training, validation, and test sets, system prompt, target modules, number of training epochs, and deterministic inference parameters. The analysis addresses training convergence, automatic test-set evaluation, bootstrap robustness, model size, parameter efficiency, and structured instructional output.

\subsection{Training Dynamics and Validation-Set Convergence}\label{training-dynamics-and-validation-set-convergence}

To compare training behavior across model sizes and LoRA ranks, we recorded the final training loss, best validation loss, validation-token accuracy, validation entropy, and best checkpoint for each fine-tuned model. Base-model parameters remained frozen, and only the LoRA adapter parameters were updated. Table~\ref{tab:5-1} reports the main training and validation results from the six experiments.

\begin{table}[t]
\centering
\caption{Training and validation results across model sizes and LoRA ranks}
\label{tab:5-1}
\scriptsize
\resizebox{\textwidth}{!}{%
\begin{tabular}{@{}llllllll@{}}
\toprule
\textbf{Model size} & \textbf{LoRA rank} & \textbf{Adapter parameters (M)} & \textbf{Training loss} & \textbf{Best validation loss} & \textbf{Validation-token accuracy (\%)} & \textbf{Validation entropy} & \textbf{Best step} \\
\midrule
3B & 4 & 1.843 & 0.8552 & 0.8830 & 78.09 & 0.8457 & 170 \\
3B & 8 & 3.686 & 0.7826 & 0.8269 & 79.03 & 0.7733 & 170 \\
3B & 16 & 7.373 & 0.7239 & 0.8080 & 79.36 & 0.7465 & 170 \\
7B & 4 & 2.523 & 0.8156 & 0.7992 & 79.57 & 0.7557 & 170 \\
7B & 8 & 5.046 & 0.7403 & 0.7599 & 80.31 & 0.7135 & 170 \\
7B & 16 & 10.093 & 0.6800 & 0.7330 & 80.71 & 0.6865 & 170 \\
\bottomrule
\end{tabular}%
}
\end{table}

Table~\ref{tab:5-1} shows that increasing the LoRA rank corresponded to lower training and validation losses for both model sizes. The best validation loss of the 3B model decreased from 0.8830 at $r=4$ to 0.8080 at $r=16$. For the 7B model, it decreased from 0.7992 at $r=4$ to 0.7330 at $r=16$. Validation-token accuracy increased with rank, whereas validation entropy decreased. Higher-rank configurations therefore produced more stable predictions of reference-answer tokens.

At the same rank, the 7B model consistently achieved a lower validation loss than the 3B model. At $r=16$, for example, the validation loss was 0.7330 for the 7B model and 0.8080 for the 3B model. Under the current data and training settings, the larger model therefore fitted the distribution of course answers more closely. This result reflects validation-set fit and should not be equated with improved domain-specific reasoning.

\begin{figure}[t]
    \centering
    \includegraphics[width=\linewidth]{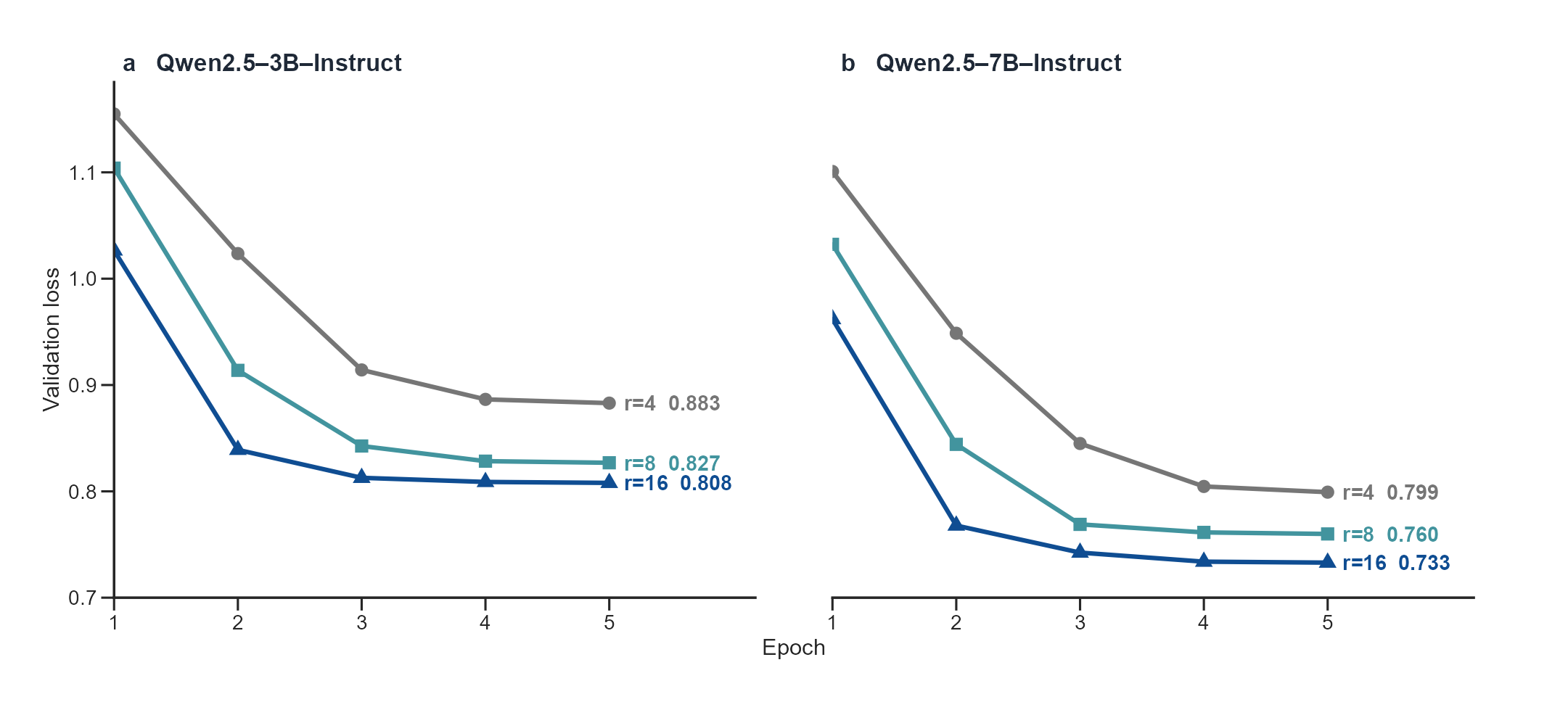}
    \caption{Validation loss across model sizes and LoRA ranks.}
    \label{fig:5-1}
\end{figure}

Fig.~\ref{fig:5-1} visualizes validation-loss convergence. Validation loss decreased with each epoch in all six experiments, with no clear rebound. Most of the decrease occurred during the first three epochs, after which gains became smaller. The models therefore reduced reference-token prediction error rapidly during early training, while later epochs provided only incremental refinement.

Higher LoRA ranks produced lower validation losses at both model sizes, indicating that greater adaptation capacity improved the fit to the distribution of course reference answers.

\subsection{Automatic Evaluation on the Test Set}\label{automatic-evaluation-on-the-test-set}

\begin{figure}[t]
    \centering
    \includegraphics[width=\linewidth]{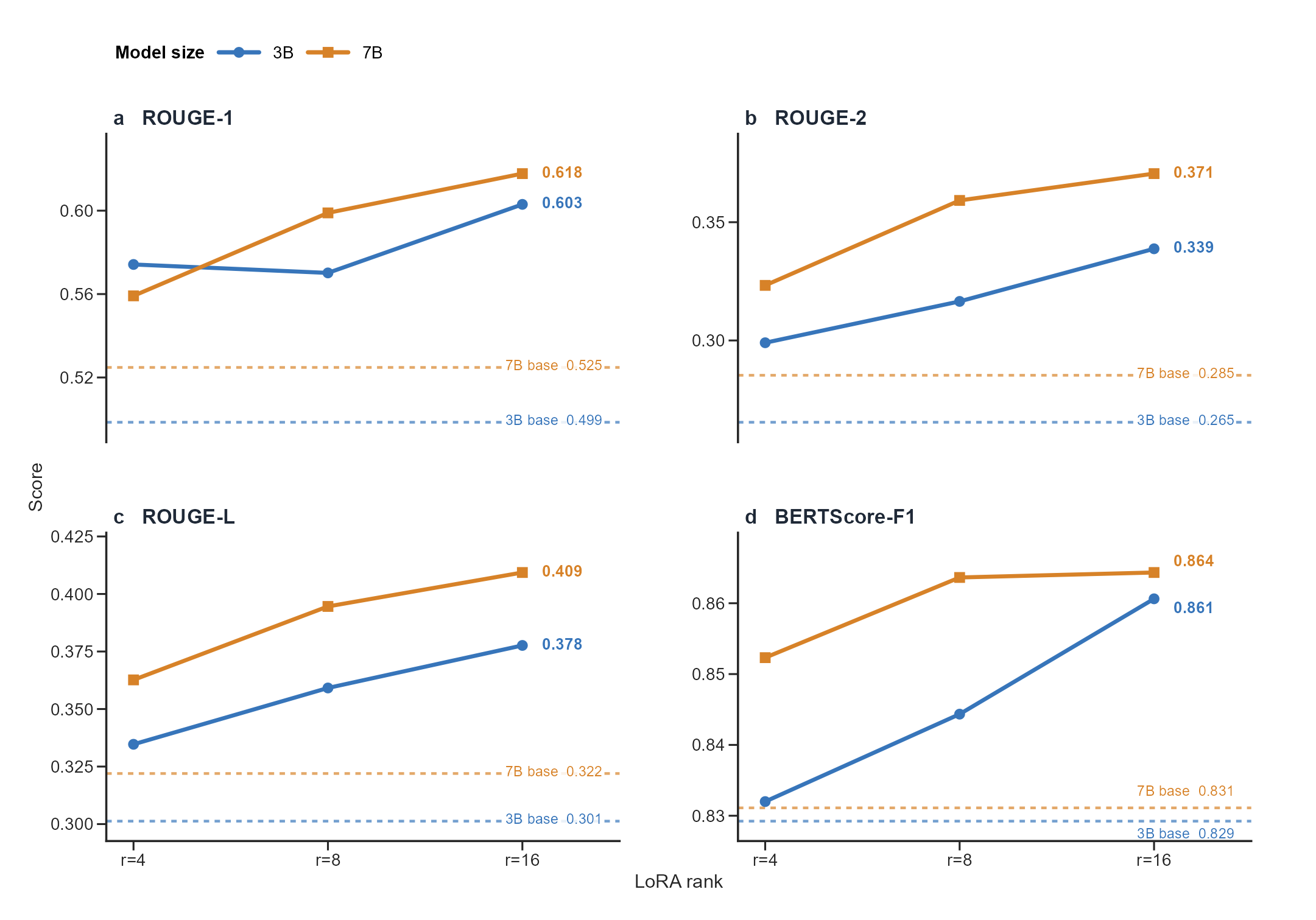}
    \caption{Automatic evaluation metrics for the base and LoRA-fine-tuned models. (a) ROUGE-1; (b) ROUGE-2; (c) ROUGE-L; (d) BERTScore-F1}
    \label{fig:5-2}
\end{figure}

We compared the base and LoRA-fine-tuned models on 54 test samples to assess how fine-tuning affected generated responses. All models used the same test set, system prompt, and deterministic generation parameters. Fig.~\ref{fig:5-2} reports ROUGE-1, ROUGE-2, ROUGE-L, and BERTScore-F1.

The LoRA-fine-tuned models outperformed their corresponding base models on most metrics. Fine-tuning on course data therefore increased both lexical overlap and semantic proximity to the reference answers. Across model sizes, the 7B models outperformed the 3B models at most ranks and were particularly consistent on ROUGE-2, ROUGE-L, and BERTScore-F1. The larger base model thus showed stronger response generation and semantic alignment on this task.

At both model sizes, $r=16$ achieved the highest ROUGE-L. ROUGE-2 and ROUGE-L generally increased with rank. ROUGE-1 also improved overall, although the 3B model showed a small fluctuation between $r=4$ and $r=8$. Increasing rank did not therefore produce monotonic gains on every metric. The difference in BERTScore-F1 between 7B-r8 and 7B-r16 was small, indicating limited marginal gains in semantic similarity from further increasing rank in the larger model. The 3B model continued to improve at $r=16$, suggesting that a smaller model may require greater adaptation capacity to fit the target representation more fully.

These automatic metrics have clear interpretive limits. As explained in Section~\ref{inference-generation-and-evaluation-metrics}, ROUGE and BERTScore measure textual and semantic proximity, not the correctness of domain-specific reasoning.

\subsection{Bootstrap Confidence Intervals and Metric Robustness}\label{bootstrap-confidence-intervals-and-metric-robustness}

We used bootstrap confidence intervals for the automatic and structured-output metrics to determine whether the test-set results were driven by a small number of samples.

\begin{figure}[t]
    \centering
    \includegraphics[width=\linewidth]{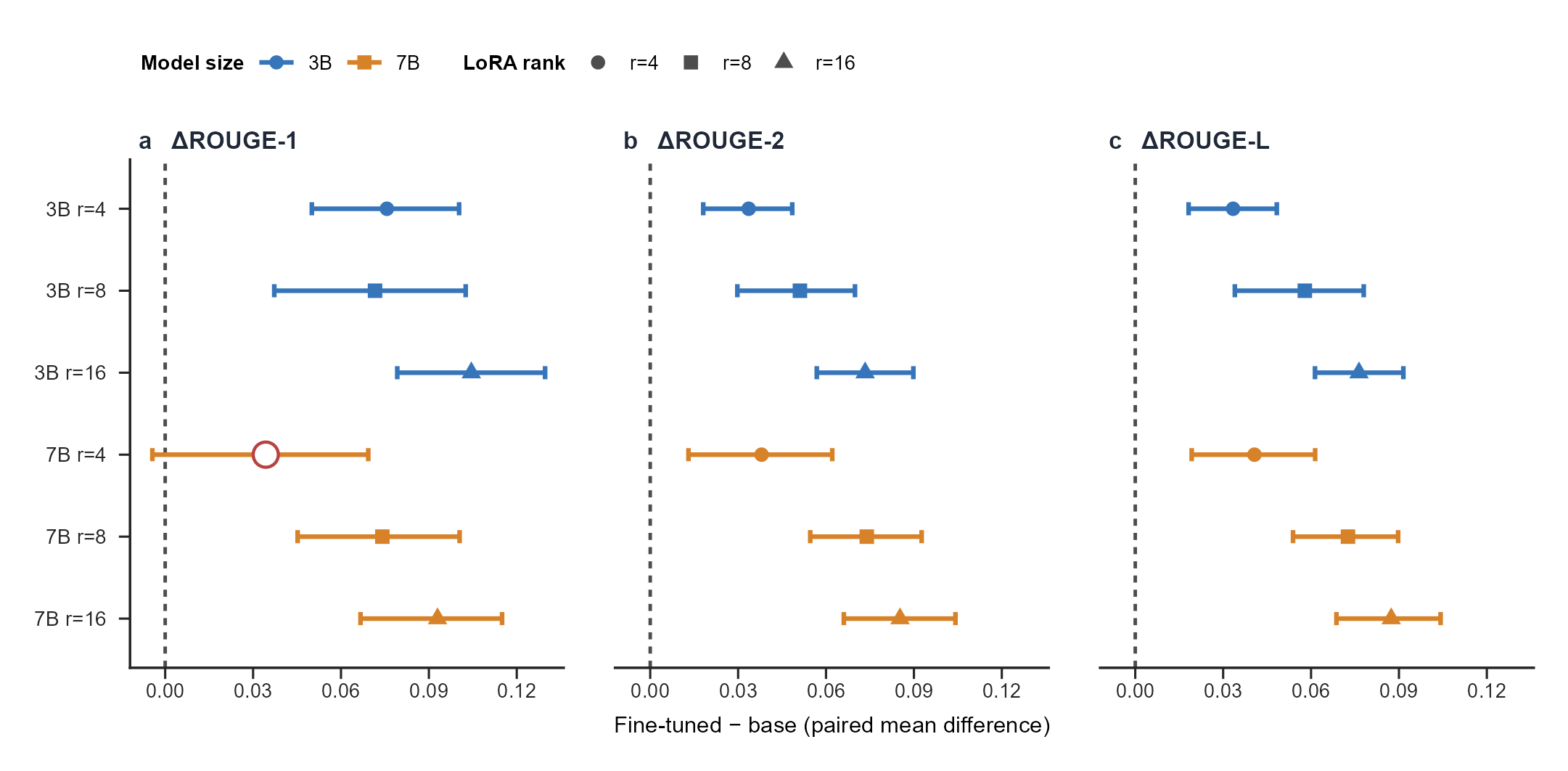}
    \caption{Bootstrap confidence intervals for ROUGE improvements.}
    \label{fig:5-3}
\end{figure}

Fig.~\ref{fig:5-3} shows the bootstrap 95\% confidence intervals for ROUGE improvements at each model size and LoRA rank. Most intervals lay entirely above zero, indicating that the higher lexical similarity between fine-tuned responses and reference answers was not attributable solely to a few test samples. For the 3B model, the confidence intervals for ROUGE-1, ROUGE-2, and ROUGE-L were above zero at $r=4$, $r=8$, and $r=16$. The ROUGE-L gain was particularly large for 3B-r16. For the 7B model, the intervals for all three ROUGE metrics were also above zero at $r=8$ and $r=16$. In contrast, the confidence interval for $\Delta$ROUGE-1 crossed zero at 7B-r4. The ROUGE-1 increase at that setting should therefore be interpreted as a trend and was less stable than the improvements under the other settings.

\begin{figure}[t]
    \centering
    \includegraphics[width=\linewidth]{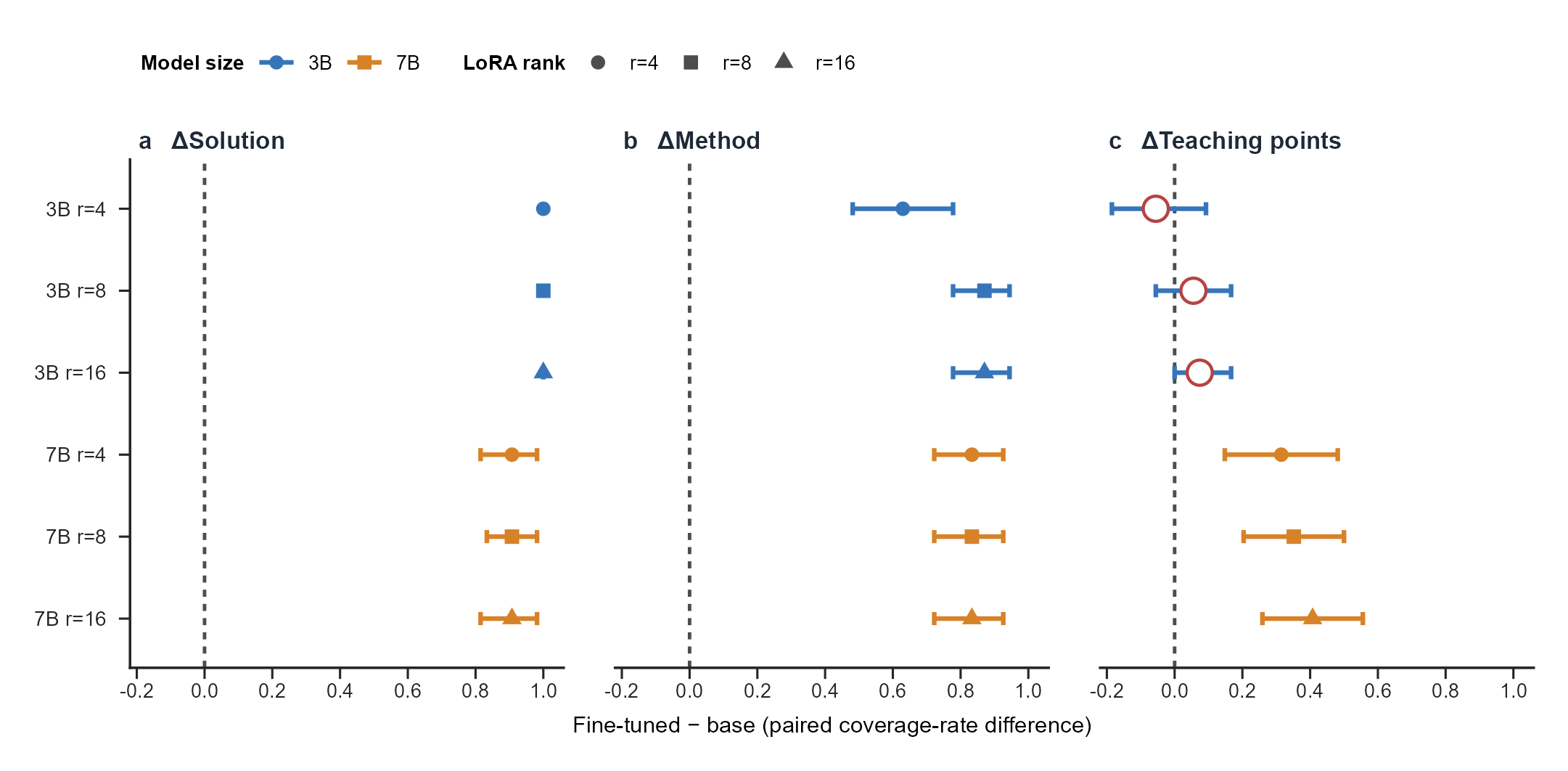}
    \caption{Bootstrap confidence intervals for improvements in structured-output coverage.}
    \label{fig:5-4}
\end{figure}

Fig.~\ref{fig:5-4} shows the gains in structured-output coverage for each LoRA-fine-tuned model relative to its corresponding base model, together with paired bootstrap 95\% confidence intervals. Solution and Method coverage improved consistently under all LoRA settings, indicating that fine-tuned models were more likely to organize their responses in the prescribed instructional format. This result is consistent with the Solution-Method-Teaching Points template used throughout the training data and shows that LoRA reinforced structured-output behavior. Improvements in Teaching Points coverage varied with model size. Some 3B settings had wider confidence intervals and lower stability, whereas improvements were more pronounced for the 7B models. Larger models may therefore learn teaching-point expressions more consistently under these conditions.

Bootstrap confidence intervals assess the stability of measured gains under changes in test-sample composition. They do not alter the evaluation boundaries described in Section~\ref{inference-generation-and-evaluation-metrics}. ROUGE and structured-output metrics reflect textual similarity and formatting stability, not the correctness of domain-specific reasoning.

\subsection{Effects of Model Size on Adaptation}\label{effects-of-model-size-on-adaptation}

Under identical data and training settings, the 7B model achieved higher mean values than the 3B model on the current test set. A direct comparison of the base models and the $r=16$ fine-tuned models illustrates how model size affected final adaptation performance.

Without course-specific adaptation, the 7B model outperformed the 3B model on all four metrics, but the margins were small. The ROUGE-L difference was +0.0207, and the BERTScore-F1 difference was only +0.0019. Thus, before adaptation, the two model sizes differed little in their surface similarity to the reference answers.

After LoRA fine-tuning at $r=16$, the size gap changed across metrics. The ROUGE-1 difference decreased from +0.0263 to +0.0148, indicating that the 3B model narrowed the lexical-overlap gap through LoRA adaptation. In contrast, the ROUGE-2 difference increased from +0.0199 to +0.0318, and the ROUGE-L difference increased from +0.0207 to +0.0317. The 7B model therefore gained more from the same LoRA setting on phrase-level matching, measured by bigram overlap, and sequence-level matching, measured by the longest common subsequence. LoRA fine-tuning did not eliminate the effect of model size. Instead, the advantage of the larger model increased on metrics that require more complex alignment of expression.

The cross-size performance of 3B-r16 is also notable. Its ROUGE-L score of 0.3776 exceeded the 0.3615 achieved by 7B-r4. Its BERTScore-F1 of 0.8606 likewise exceeded the 0.8579 of 7B-r4. A smaller model with higher-rank LoRA adaptation can therefore outperform a larger model with lower-rank adaptation on some metrics. This finding has direct practical relevance for instructional applications that require local deployment or operate under limited resources.

\subsection{LoRA Rank and Parameter Efficiency}\label{lora-rank-and-parameter-efficiency}

LoRA rank affects both metric performance and the number of adapter parameters. To avoid selecting a model solely by its highest score, we compared parameter efficiency across ranks. Table~\ref{tab:5-3} reports the ROUGE-L and BERTScore-F1 gains of each fine-tuned model over its corresponding base model. It also reports the ROUGE-L gain per million adapter parameters.

We define parameter efficiency as the ROUGE-L improvement per million adapter parameters:

\[
E=\frac{\Delta\mathrm{ROUGE\text{-}L}}{P_{\mathrm{adapter}}/10^{6}}
\]

Here, \((P_{adapter})\) is the number of LoRA adapter parameters, and \((E)\) is the ROUGE-L improvement per million adapter parameters. The absolute improvement is:

\[
\Delta\mathrm{ROUGE\text{-}L}=\mathrm{ROUGE\text{-}L}_{\mathrm{LoRA}}-\mathrm{ROUGE\text{-}L}_{\mathrm{Base}}
\]

This metric quantifies the relative return between parameter cost and performance improvement at different LoRA ranks.

\begin{table}[t]
\centering
\caption{Comparison of LoRA rank and parameter efficiency}
\label{tab:5-3}
\scriptsize
\resizebox{\textwidth}{!}{%
\begin{tabular}{@{}llllll@{}}
\toprule
\textbf{Model size} & \textbf{LoRA rank} & \textbf{Adapter parameters (M)} & \textbf{$\Delta$ROUGE-L} & \textbf{$\Delta$BERTScore-F1} & \textbf{$\Delta$ROUGE-L per million parameters} \\
\midrule
3B & 4 & 1.843 & +0.0334 & +0.0028 & 0.0181 \\
3B & 8 & 3.686 & +0.0579 & +0.0151 & 0.0157 \\
3B & 16 & 7.373 & +0.0764 & +0.0314 & 0.0104 \\
7B & 4 & 2.523 & +0.0407 & +0.0212 & 0.0161 \\
7B & 8 & 5.046 & +0.0726 & +0.0325 & 0.0144 \\
7B & 16 & 10.093 & +0.0874 & +0.0332 & 0.0087 \\
\bottomrule
\end{tabular}%
}
\end{table}

Table~\ref{tab:5-3} shows that $r=16$ produced the largest absolute ROUGE-L gain at both model sizes but was less parameter-efficient than $r=4$ and $r=8$. For the 7B model, the $\Delta$ROUGE-L was +0.0726 at $r=8$, corresponding to a gain of 0.0144 per million adapter parameters, whereas $r=16$ increased the absolute gain to +0.0874 but reduced parameter efficiency to 0.0087. A similar pattern occurred for the 3B model. These results show that increasing LoRA rank improved absolute performance but produced diminishing returns in performance per trainable parameter.

\begin{figure}[t]
    \centering
    \includegraphics[width=\linewidth]{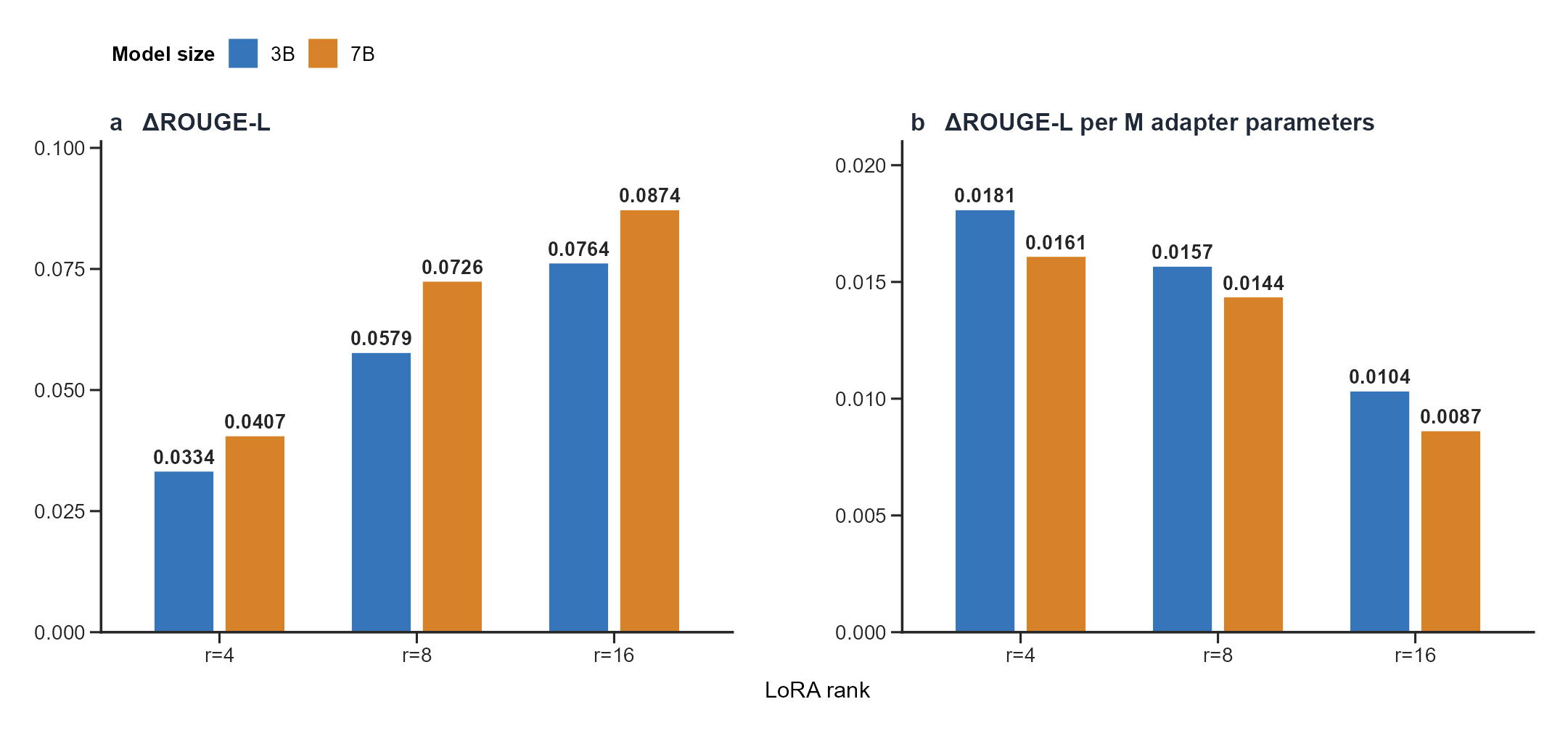}
    \caption{Relationship among LoRA rank, absolute ROUGE-L improvement, and parameter efficiency. Although $r=16$ achieved the largest absolute gain, the ROUGE-L gain per adapter parameter decreased as rank increased.}
    \label{fig:5-5}
\end{figure}

In practical terms, $r=4$ required the fewest parameters, but produced only a small BERTScore-F1 gain for the 3B model and less complete structured output. The $r=8$ setting provided a stable compromise between parameter count and performance. The $r=16$ setting is more suitable when maximizing reference-answer similarity and structural completeness is the primary objective. Its diminishing returns should nevertheless be considered together with training cost and overfitting risk. Fig.~\ref{fig:5-5} visualizes this trade-off between absolute performance and parameter efficiency.

\subsection{Analysis of Structured Instructional Output}\label{analysis-of-structured-instructional-output}

\begin{figure}[t]
    \centering
    \includegraphics[width=\linewidth]{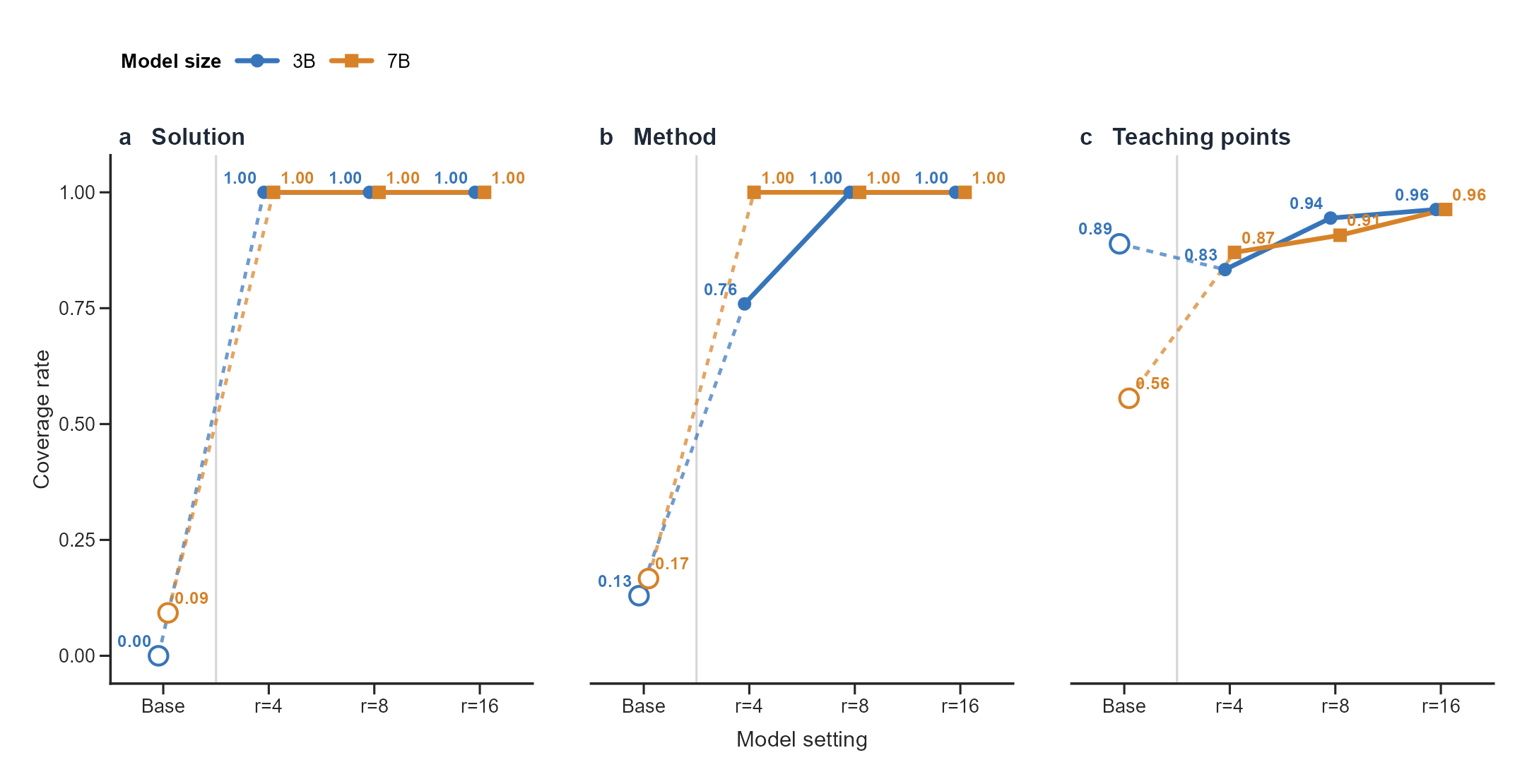}
    \caption{Coverage of structured instructional output in the base and LoRA-fine-tuned models.}
    \label{fig:5-6}
\end{figure}

In addition to textual and semantic similarity, we examined whether the models generated responses with the structure expected in instructional question answering. All assistant responses in the dataset use the Solution-Method-Teaching Points format. We therefore measured the coverage of the Solution, Method, and Teaching Points markers to determine whether the models learned the pedagogical template in the training data.

Fig.~\ref{fig:5-6} shows that LoRA fine-tuning substantially increased coverage of the Solution and Method markers. For the 3B base model, Solution coverage was 0 and Method coverage was only 0.13. Both reached 1.00 at $r=8$ and $r=16$. For the 7B base model, Solution and Method coverage were 0.09 and 0.17, respectively, whereas both reached 1.00 under all three LoRA settings. Fine-tuning on the course data therefore made the models more likely to follow the prescribed instructional structure and substantially improved structured-output stability.

Changes in Teaching Points coverage were more complex. The 3B base model already had relatively high coverage, but its Solution and Method coverage was low. It could thus produce summary-like teaching hints without constructing the complete Solution-Method-Teaching Points structure. At $r=8$ and $r=16$, Teaching Points coverage increased to 0.94 and 0.96, respectively, while both Solution and Method coverage reached 1.00. Higher-rank settings therefore supported a more complete structured instructional response in the 3B model. For the 7B model, Teaching Points coverage increased from 0.56 in the base model to 0.87, 0.91, and 0.96 across the LoRA settings. The positive trend was more consistent.

Together with the bootstrap results in Section~\ref{bootstrap-confidence-intervals-and-metric-robustness}, these findings show that improvements in Solution and Method coverage were stable under most LoRA settings. Improvements in Teaching Points coverage were more sensitive to model size and rank. Overall, LoRA fine-tuning increased both reference-answer similarity and consistency in the pedagogical organization of responses.

\subsection{Overall Interpretation and Evaluation Boundaries}\label{overall-interpretation-and-evaluation-boundaries}

The analyses above support three main findings.

First, LoRA fine-tuning brought responses from both model sizes closer to the course reference answers at the lexical and semantic levels. Improvements in ROUGE-L and structured-output metrics were stable under bootstrap resampling.

Second, model size and LoRA rank jointly affected adaptation. The 7B model generally outperformed the 3B model, and $r=16$ achieved the highest absolute scores. The $r=8$ setting offered a more conservative compromise between performance and parameter efficiency.

Third, all experimental groups showed stable gains in Solution and Method coverage, whereas stable gains in Teaching Points coverage were concentrated in the 7B group. Neither the automatic metrics nor the structured-output metrics directly assess the correctness of domain-specific reasoning. Section~\ref{interpretive-boundaries-of-the-evaluation-metrics} discusses this boundary further.

\section{Discussion}\label{discussion}

\subsection{What Did LoRA Fine-Tuning Improve?}\label{what-did-lora-fine-tuning-improve}

The results in Section~\ref{experimental-results-and-analysis} show that LoRA fine-tuning produced its clearest improvements in reference-answer similarity and the stability of structured instructional output. This section discusses the nature and educational implications of these changes.

Greater reference-answer similarity indicates that the models more closely followed the vocabulary, formula presentation, and step organization of the standard-answer patterns in the course data. Pedagogically, this alignment may reduce students' cognitive load because the generated solutions resemble the style of the textbook and classroom explanations instead of varying unpredictably. However, stylistic adaptation does not ensure content accuracy. A LoRA-fine-tuned model may produce a complete-looking derivation in the expected format while making an error in symbol substitution or algebra. Reference-answer similarity is therefore better understood as formal alignment with the course than as substantive improvement in reasoning ability.

The increased stability of structured output was particularly evident in Solution and Method coverage, which rose from near zero to 100\%. LoRA fine-tuning therefore internalized the response template in the training data as a consistent generation pattern. This behavior has practical value in instructional deployment. If a model reliably organizes content as Solution, Method, and Teaching Points, instructors and students can anticipate the basic response format, making answers easier to review, verify, and compare. Improvements in Teaching Points coverage varied with model size. The 3B base model already exhibited some coverage, whereas the 7B model increased from 0.56 to 0.96. This difference suggests that generation of teaching-point content may be more sensitive to model size.

The principal value of LoRA fine-tuning in this study was thus its transfer of course-specific language and response organization into model outputs. The models shifted from being able to answer control systems questions to answering them in the manner expected by the course. This improvement concerns form and style and should not be equated with better domain-specific reasoning.

\subsection{Trade-Off Between Model Size and LoRA Rank}\label{trade-off-between-model-size-and-lora-rank}

Our experimental design varied both model size and LoRA rank, allowing us to examine their joint relationship rather than reporting each factor in isolation.

The 7B model outperformed the 3B model on almost every metric, which is consistent with the general expectation that larger models have stronger language-modeling capabilities. More importantly, the size gap changed across conditions. Before fine-tuning, the 7B model's ROUGE-L advantage over the 3B model was +0.0207. After LoRA fine-tuning at $r=16$, the advantage increased to +0.0317. LoRA therefore did not eliminate the difference between model sizes. Under the same rank setting, the larger model achieved a greater absolute gain. Under the conditions examined, adapting a larger base model with LoRA may be preferable to repeatedly increasing rank on the same smaller model.

At both model sizes, $r=16$ achieved the highest absolute metric values. Parameter efficiency, measured as ROUGE-L gain per million adapter parameters, nevertheless decreased continuously from $r=4$ to $r=16$. For the 7B model, parameter efficiency was 0.0161 at $r=4$, 0.0144 at $r=8$, and 0.0087 at $r=16$, showing clear diminishing marginal returns. The same trend occurred in the 3B model. The $r=8$ setting thus offered a relatively robust compromise in these experiments. It substantially outperformed $r=4$ while using approximately half as many adapter parameters as $r=16$.

These results support three configuration choices for the settings studied here. First, when the objective is to maximize reference-answer similarity without strict resource constraints, such as in cloud deployment, 7B-r16 is the strongest configuration. Second, for local deployment or use on a consumer-grade GPU, 3B-r8 or 7B-r8 provides a favorable balance between parameter count and performance. Third, when the goal is rapid prototyping or an initial feasibility test, $r=4$ on the 7B model still yields a meaningful improvement (\(\Delta\)ROUGE-L = +0.0407).

\subsection{Interpretive Boundaries of the Evaluation Metrics}\label{interpretive-boundaries-of-the-evaluation-metrics}

We used ROUGE, BERTScore, and structured-output metrics because they capture different aspects of similarity to the reference answers \cite{ref28,ref29,ref30,ref31,ref32,ref35,ref36}. ROUGE measures lexical overlap, BERTScore-F1 measures semantic proximity, and the structured-output metrics indicate whether a model consistently generates pedagogical components such as Solution, Method, and Teaching Points. Together, these metrics support the conclusion that LoRA improved alignment with the reference answers and their organizational format.

The limitations of these metrics are equally clear. ROUGE is sensitive to wording and surface overlap. BERTScore does not verify mathematical symbols or formula derivations. Structural metrics based on headings or keywords cannot determine whether a derivation is valid. A response may contain every required heading while including incorrect formulas, inadequate variable explanations, or an incomplete conclusion. Clear structure and reference-answer similarity do not therefore establish domain-specific correctness.

This distinction is especially important in engineering education. A course question-answering model ultimately serves student learning, and instructional reliability depends on correct formulas, rigorous derivations, explicit assumptions, and clear explanations. The present framework is suitable for measuring textual similarity and structural stability, but it cannot replace evaluations of answer quality by instructors or domain experts.

\subsection{Limitations}\label{limitations}

First, the study is limited by dataset size and coverage. The dataset contains 360 samples, including 54 test samples. It supports an initial feasibility analysis but does not cover all topics, question types, or levels of difficulty in a control systems course. Complex controller design, integrated modeling, multistep state-space analysis, and open-ended explanatory questions require more samples for reliable evaluation.

Second, the range of models is limited. We compared Qwen2.5-3B-Instruct and Qwen2.5-7B-Instruct but did not include other open-source model families, such as Llama or DeepSeek \cite{ref39,ref40}, or models with more parameters. The conclusions are therefore restricted to the current Qwen2.5 models, dataset, and LoRA configurations and should not be generalized directly to all LLMs.

Third, evaluation relies primarily on automatic metrics and rule-based detection of structural markers. These measures quantify reference-answer similarity and formatting stability but do not adequately assess formula accuracy, derivational rigor, or instructional usefulness. Future work should involve control systems instructors or domain experts and use a human-evaluation rubric covering correctness, completeness, derivational quality, and instructional clarity.

Fourth, experimental robustness can be improved. We used bootstrap resampling to report 95\% confidence intervals for test-set ROUGE and structured-output metrics, but BERTScore-F1 is still reported only as a mean. The experiments also used a single training seed and do not report means and variances across multiple seeds. In addition, no strong-prompt or RAG baseline was included. The current design therefore cannot fully separate the effects of fine-tuning, prompt design, and external knowledge retrieval.

\section{Conclusions and Future Work}\label{conclusions-and-future-work}

\subsection{Conclusions}\label{conclusions}

We constructed a structured dataset of 360 system-user-assistant dialogues for question answering in a control systems course. Using Qwen2.5-3B-Instruct and Qwen2.5-7B-Instruct as base models, we compared LoRA ranks of $r=4$, $r=8$, and $r=16$ under a unified experimental protocol. The main conclusions are as follows.

First, LoRA fine-tuning improved reference-answer similarity and structured-output stability at both model sizes. The best configuration, 7B-r16, achieved a ROUGE-L of 0.4093 and a BERTScore-F1 of 0.8643. Bootstrap analysis yielded 95\% confidence intervals of {[}0.0613, 0.0915{]} and {[}0.0687, 0.1042{]} for the ROUGE-L gains of 3B-r16 and 7B-r16, respectively. Neither interval crossed zero, indicating that both gains were robust to bootstrap resampling.

Second, model size and LoRA rank had joint effects on adaptation. The larger model achieved a greater absolute benefit from the same adaptation setting. After fine-tuning, the 7B model's ROUGE-L advantage over the 3B model increased from +0.0207 to +0.0317. However, 3B-r16 outperformed 7B-r4 on some metrics, indicating that higher-rank adaptation can partially compensate for smaller model size. The marginal return of rank decreased as rank increased. The $r=8$ setting approached the performance of $r=16$ with approximately half as many adapter parameters, making it a robust compromise for resource-constrained applications.

Third, the evaluation framework focuses on reference-answer similarity and the stability of pedagogical structure. ROUGE, BERTScore, and the structured-output metrics do not directly assess the correctness of domain-specific reasoning. These findings should therefore be understood as evidence of improved course alignment in form and style, not as certification of domain-specific reasoning. Formula accuracy, derivational rigor, and instructional reliability still require expert evaluation.

Overall, LoRA fine-tuning provides a feasible and parameter-efficient method for aligning open-source instruction-tuned models with the response style and structured-output requirements of a specific engineering course. Model size and LoRA rank should be selected jointly according to deployment resources, performance objectives, and the amount of available data.

\subsection{Future Work}\label{future-work}

Future research should first expand the control systems question-answering dataset to cover more topics, question types, and difficulty levels. Additional samples are particularly needed for complex system modeling, frequency-domain analysis, state-space design, controller design, and integrated derivation tasks. Reference-answer quality should also be improved through greater consistency in formulas, variables, assumptions, and pedagogical explanations.

Second, evaluation should extend beyond automatic text metrics and structural-marker detection. Building on the current bootstrap analysis, future studies could use multiple random seeds, expert-evaluation rubrics, and stronger baselines. Model outputs could then be assessed for formula accuracy, derivational rigor, answer completeness, explanatory clarity, and instructional usability \cite{ref30,ref31,ref32,ref35,ref36}.

Third, future experiments should include more models and baselines \cite{ref5,ref6,ref17,ref39,ref40}. Other open-source model families, including Llama and DeepSeek, could be compared at the model level. At the method level, the comparison could include strong-prompt baselines, RAG baselines, and combined LoRA-RAG systems. These experiments would help distinguish the respective contributions of model size, prompt design, external knowledge retrieval, and parameter-efficient fine-tuning.

Finally, deployment strategies for instructional settings warrant further investigation. LoRA fine-tuning helps models learn the language and organization of course answers, whereas RAG can provide external evidence from textbooks, lecture notes, and formula sheets \cite{ref13,ref17}. Combining the two approaches may improve course relevance, traceability, and structural stability. Their educational value, however, must still be validated by domain experts and in authentic teaching settings.

\end{document}